\documentclass[runningheads]{llncs2}
\usepackage{graphicx}
\usepackage{amsmath} 
\usepackage{amssymb}  
\usepackage{hyperref}
\usepackage[table]{xcolor}
\usepackage{float}
\usepackage{setspace}
\usepackage{fancyhdr}
\usepackage{multirow}
\usepackage{caption}
\usepackage{subcaption}
\usepackage{dirtytalk}

\begin{document}

\title{Dissecting FLOPs along input dimensions for GreenAI cost estimations
}


\author{Andrea Asperti\inst{1} \and Davide Evangelista\inst{2} 
\and Moreno Marzolla\inst{1}}

\institute{
University  of Bologna\\
Department of Informatics: Science and Engineering (DISI)\\
\and
University  of Bologna\\
Department of Mathematics\\
}


\maketitle

\begin{abstract}
The term GreenAI refers to a novel approach to Deep Learning, 
that is more aware of the ecological impact and the computational 
efficiency of its methods. The promoters of GreenAI suggested the use of Floating Point Operations (FLOPs) as a measure of
the computational cost of Neural Networks; however, that measure 
does not correlate well with the energy consumption of hardware 
equipped with massively parallel processing units like GPUs or TPUs.
In this article, we propose a simple refinement of the formula used 
to compute floating point operations for convolutional layers, called $\alpha$-FLOPs, explaining and correcting the traditional discrepancy with respect to different layers, and closer to reality. The notion of $\alpha$-FLOPs 
relies on the crucial insight that, in case of inputs with multiple 
dimensions, there is no reason to believe that the speedup offered by 
parallelism will be uniform along all different axes.
\end{abstract}

\section{Introduction}
Artificial Intelligence, especially in its modern incarnation
of Deep Learning, has achieved remarkable results in 
recent years, matching -- and frequently trespassing -- human capabilities in a number of different tasks. These
techniques usually require the deployment of massive computational resources, with huge implications in terms 
of energy consumption. To make a couple of examples the hyper-realistic Generative Adversarial Network
for face generation in~\cite{InvidiaGAN18} required training on 8 Tesla V100 GPUs 
for 4 days; the training of BERT~\cite{BERT}, a well known generative model for NLP, takes about 96 hours on~64 TPU2 chips.
Researchers at the University of Massachusetts~\cite{5cars} have recently
performed a life cycle assessment relative to the training of large state-of-the-art AI models, 
discovering that the process can emit a quantity of carbon dioxide roughly equivalent to the lifetime
emissions of five medium cars. Other authors reached similar conclusions~\cite{lacoste2019quantifying}.

Until a few years ago, the ecological impact of artificial intelligence was entirely neglected by 
researchers and industry, who were mostly focused on improving performance at any cost. However, this 
has changed in recent years, with a growing awareness that this trend of research is not sustainable 
any more~\cite{AIsustainability}, and an increased attention towards energetic 
efficiency~\cite{efficientNet}.

The GreenAI paper~\cite{GreenAI} summarizes well the goal and objectives of the new 
philosophy: it promotes a new practice in Deep Learning, that is more focused on the social costs of 
training and running models~\cite{carbontracker,measurement,ChasingCO2}, encouraging the 
investigation of increasingly efficient models~\cite{EfficientTransformers,VAEsurvey}. 

To this aim, it is essential to identify widely acceptable and reliable metrics to assess and compare
the cost and efficiency of different models. Several metrics are investigated and discussed
in~\cite{GreenAI}; in conclusion, the number of Floating Point Operations (FLOPs) is
advocated and promoted, since it is easily computed for Neural Networks while offering a hardware 
independent, schematic but meaningful indication of the actual computation cost of 
the model~\cite{lacoste2019quantifying}.  

Unfortunately, the mere computation of FLOPs does not cope well with the massively parallel 
architectures (GPU and TPU) typically used in Deep Learning~\cite{trap_of_flops}. Efficient 
implementation of neural networks on these architectures depends both on 
complex algorithms for General Matrix Multiplication
(GEMM)~\cite{GEMM2015} and sophisticated load balancing techniques~\cite{loadbalancingGPUs} 
splitting the workload on the different execution units. As we shall see, these algorithms usually 
perform better for specific layers and, especially, {\em along specific axes} 
of the input dimension of these layers. 

Our claim is that it is possible to study the performance of neural layers 
(especially, convolutions) as \say{black boxes}, measuring the execution time for
a number of different configurations, and separately investigating 
the execution time for increasing dimensions along different axis.

As a result, we propose a simple correction to the formula used to compute FLOPs for convolutional layers, 
that provides better estimations of their actual cost, and helps to understand the discrepancy with respect to the cost of different layers. 

\paragraph{Organization of the article} This paper has the following
structure. In Section~\ref{sec:measures} we briefly discuss some possible metrics for measuring the 
efficiency of models; we particularly focus on FLOPs, discussing their computation 
for some basic operations relevant for Neural Networks.
In Section~\ref{sec:comp-flops} we introduce the GEMM (GEneral Matrix Multiply) operation, that helps to understand the canonical computation of FLOPs for the Convolution layers. 
In Section~\ref{sec:problem_of_convolution} we present some experiments which show that, if Convolutions are 
executed on GPU, FLOPs are not a good measure for efficiency. That is the motivation for introducing a 
correction, that we call $\alpha$-FLOPs, defined and discussed in Section~\ref{sec:alpha_flops}.
Section~\ref{sec:more_experiments} offers more experimental results,
validating the formula with respect to growing input dimensions along specific axes.

\section{Measures of Efficiency}\label{sec:measures}

In this section we review some of the metrics that can be used to measure the efficiency of an~AI algorithm, following the discussion of~\cite{GreenAI}.

\paragraph{Carbon Emission}
As already remarked in the introduction, the present work is motivated by the need to reduce the energy 
consumption of training large state-of-the-art AI models. Unless a significant fraction of such energy comes 
from renewable sources, reducing the power required for AI training means that less carbon dioxide is released into the atmosphere. Unfortunately, precise quantification of carbon emission associated with computational 
tasks is impractical, since it depends both on the hardware hosting the computation, and also on the local 
energy production and distribution infrastructure.

\paragraph{Number of parameters}
The number of parameters of a Deep Learning model is an interesting and hardware-independent measure of the 
complexity of models. Unfortunately, the number of parameters alone is poorly correlated with the total 
training time, since parameters may refer to different operations. For example, Convolutional Layers have 
relatively few parameters, relative to the kernel of the convolution; this does not take into account the 
actual cost of convolving the kernel over the input.

\paragraph{Execution time}
The total running time is a natural measure of efficiency: faster algorithms are better. 
Execution time depends on the number of instructions executed and hence is strictly 
correlated with the total energy consumption~\cite{perf-counter-power-consumption}; therefore, 
it is a good proxy of power usage when direct energy measurement is impractical.
There are a couple of important considerations that must be made when considering execution time 
as a metric: (\emph{i})~it requires an implementation of the algorithm being measured, 
which may take time and effort to be developed; (\emph{ii})~execution time is hardware- 
and language-dependent, since it depends on both the underlying hardware and on the efficiency of 
the compiler/interpreter. 

\paragraph{FLOPs}
The number of FLoating Point OPerations (FLOPs) is a metric that is widely used in the context of
numerical computations~\cite{comparch,morphnet,super_efficient_networks,pruning-CNN}. It
is defined as the total count of elementary machine operations (floating point additions and 
multiplications) executed by a program. Floating point operations have a latency of several CPU 
cycles on most current processor architectures~\cite{latency-intel,latency-amd,latency-arm}, although
the use of pipelining, multiple-issue and~SIMD instructions significantly increase the throughput. In
general, floating point operations have higher latency than most of the other CPU instructions (apart
from load/stores from/to main memory, where memory access is the bottleneck); therefore, they tend to
dominate the execution time of numerical algorithms. For this reason, the number of floating point 
operations is used as a proxy for the execution time of a program.

As an example, suppose that~$v$ and~$w$ are $n$-dimensional arrays. Then, the inner product between~$v$ and~$w$

\begin{align}\label{eq:inner_product}
    \langle v; w \rangle = \sum_{i=1}^n v_i w_i
\end{align}

requires~$n$ multiplications and~$n-1$ additions, for a total of $2n - 1$ FLOPs.
Similarly, the matrix-vector product between an $m \times n$ matrix~$A$ and an $n$-dimensional vector~$v$ requires~$m$ inner product, for a total of $2mn - m$ FLOPs. 

Since operations similar to~\eqref{eq:inner_product}, where a sequence of multiplications are added together, are very common, modern CPUs supports FMA (Fused Multiply-Add) instructions, where a multiplication followed by an addition are executed as a single operation and require less time than two separate instructions. For this reason, the definition of FLOPs is usually modified to be the total number of FMA operations needed for a full iteration of an algorithm. With this definition (that it is usually followed by some authors), the inner product of two $n$-dimensional arrays requires~$n$ FLOPs, while the product between an $m \times n$ matrix with an $n$-dimensional vector requires $nm$ FLOPs. Nonetheless, since we are interested in measuring the performance under massively parallel architectures, through this paper we will follow the classical definition of FLOPs.

\section{Computation of FLOPs for basic layers}\label{sec:comp-flops}

The basic operation that dominates training of Neural Network models is the dense matrix-matrix 
product. This operation is often referred in the technical literature as~GEMM (for \emph{GEneral 
Matrix Multiply}), owing its name to the \emph{x}\texttt{GEMM} family of functions provided by the
Basic Linear Algebra Subprograms (BLAS) library~\cite{blas-2002}. BLAS is a widely used collection
of subroutines implementing basic operations involving vectors and matrices, such
as vector addition, dot product, vector-matrix multiplication and so on; these functions act as
building blocks on which more complex linear algebra computations can be programmed. 
Being at the core of many applications, the performance of BLAS primitives
are critical, so most hardware vendors provide their own optimized implementations, 
e.g., \texttt{cuBLAS} for nVidia GPUs~\cite{cublas}, and \texttt{clBLAS} for OpenCL 
devices~\cite{clblas}, including various brands of GPUs and multicore processors.

A~GEMM operation takes the general form:

\begin{equation}
\mathbf{C} \leftarrow \alpha \mathbf{AB} + \beta \mathbf{C}\label{eq:gemm}
\end{equation}

\noindent where $\mathbf{A}, \mathbf{B}, \mathbf{C}$ are matrices of compatible size, and 
$\alpha, \beta$ are scalars. The matrix-matrix product
$\mathbf{C} \leftarrow \mathbf{AB}$ is a special case of \eqref{eq:gemm} where 
$\alpha =1$, $\beta = 0$.

Assuming that the size of $\mathbf{A}$ is $m \times k$ and the size of $\mathbf{B}$
is $k \times n$, then the size of $\mathbf{C}$ must be $m \times n$ and the direct computation of~\eqref{eq:gemm} using vector dot products requires:

\begin{itemize}
\item $2mkn + mn$ FLOPs for the matrix product $\alpha \mathbf{AB}$, assuming that
dot products are implemented with an inner loop involving a multiply-accumulate
operation like $s \leftarrow s + x_i y_i$
\item $mn$ FLOPs for the computation of $\beta \mathbf{C}$
\item $mn$ additional FLOPs for the computation of the matrix sum $\alpha \mathbf{AB} + \beta \mathbf{C}$
\end{itemize}

\noindent from which we get that a total count of $2mkn + mn + mn + mn = mn(2k+3)$ FLOPs are required
for the general~GEMM. Neglecting lower-order terms we can approximate the operation count with
$2mkn$.

We can apply this result for the layers of a Neural Network. 
Consider a Dense layer, with input and output dimensions $D_\textit{in}$ and $D_\textit{out}$,
respectively. We need to compute the product between the weight matrix of size
$D_\mathit{out} \times D_\mathit{in}$ and the input, plus a bias term $B$ of dimension 
$D_{out}$; therefore, the number of FLOPs is

\begin{equation*}
2 D_\mathit{in} D_\mathit{out} - D_\mathit{in} + D_\mathit{out} 
\end{equation*}

As above, we omit the lower order terms as they are asymptotically negligible. As a consequence, 
we will consider a Dense layer to have a number of FLOPs equal to

\begin{equation}\label{eq:dense_flops}
2 D_{in} D_{out}
\end{equation}

The case of a convolutional layer is slightly more complex. Let us consider the case of a~2D
convolution. Let $(W_\textit{in},H_\textit{in},C_\textit{in})$
the dimension of the input (written with the notation ({\em Width, Height, Channels})), 
$(W_\textit{out},H_\textit{out},C_\textit{out})$ the dimension of the output (depending on 
the stride and number of kernels), and let $K_1,K_2$ be the dimensions of the kernel. 
Then, the number of FLOPs is given by

\begin{equation}\label{eq:conv_flops}
    2 \cdot \underbrace{K_1 \cdot K_2 \cdot C_\textit{in}}_{\text{kernel dim}} \cdot \underbrace{W_\textit{out} \cdot H_\textit{out}}_{\text{input dim}}\;\ \cdot \underbrace{C_\textit{out}}_{\text{output dim}}
\end{equation}


In the following, we shall frequently consider the case of convolutions with stride~$1$ in \say{same}
padding modality. In this case, $W_\textit{in}=W_\textit{out}$ and 
$H_\textit{in}=H_\textit{out}$, so we shall drop the subscripts, and just write~$W$ and~$H$.
Moreover, in the frequent case kernels are squared, we drop the subscripts in $K_1,K_2$ and just write~$K$.


\section{The problem of convolutions}\label{sec:problem_of_convolution}

A dense layer of dimension $D_{in} \times D_{out}$ is the same as a unary convolution $(K=1)$ with 
$C_{in} = D_{in}$, $C_{out} = D_{out}$ and $H=W=1$; it is easy to experimentally check that both 
require the same time to be evaluated. However, as soon as we distribute the total number of FLOPs 
of equation~\eqref{eq:conv_flops} across different dimensions, we observe a significant speedup, that
has no justification in terms of FLOPs. This raises concerns about the use of FLOPs for estimating 
running time (and hence energy consumption). In this section we provide empirical evidence of this 
phenomenon.

\begin{figure}[ht]
\begin{subfigure}{.5\textwidth}
\begin{tabular}{|c|c|}\hline
\multicolumn{2}{|c|}{327.68 M FLOPs}\\\hline\hline
\multicolumn{2}{|c|}{\bf Dense layer}\\\hline
 ($D_1,D_2$)  & time (ms)\\\hline
$(12800,12800) $  & 6.401  \\\hline\hline
\multicolumn{2}{|c|}{\bf Convolutional layers}\\\hline
$(W,H,C_{in},C_{out},K_1,K_2)$ & time (ms)\\\hline
$(1,1,12800,12800,1,1)$ &  6.392 \\\hline
$(1,2,,6400,12800,1,1)$ &  3.224 \\\hline
$(2,2,6400,6400,1,1)$ & 1.626 \\\hline
$(4,4,3200,3200,1,1)$ & 0.454  \\\hline
\end{tabular}
\caption{\label{fig:dense_a}}
\end{subfigure}
\begin{subfigure}{.5\textwidth}
\includegraphics[width=1\textwidth]{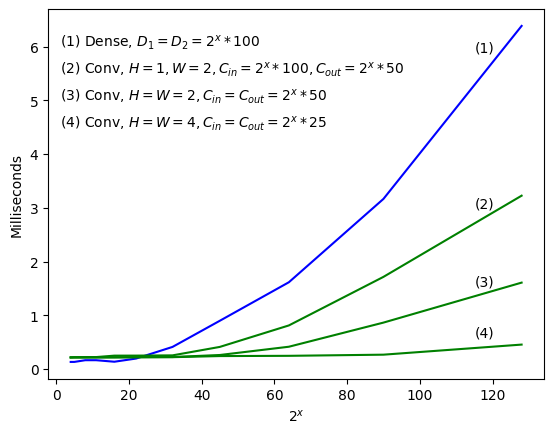}
\caption{\label{fig:dense_b}}
\end{subfigure}
\caption{\label{fig:dense_vs_conv}Comparison of execution times for Dense and Convolutional 
layers with the \emph{same} amount of FLOPs. In 
Table~(a) we provide numerical values for layers with
327.68 Million FLOPs; in the right we show the execution time
of similar configurations for increasing dimensions.
All layers for a given value of~$2^x$ (i.e. along any vertical section) have the same amount of FLOPs.}
\end{figure}

In Figure~\ref{fig:dense_vs_conv}, we compare the time required to evaluate a dense layer 
with several different convolutional layers with a same amount of FLOPs computed according 
to~\eqref{eq:conv_flops}; the execution time has been measured on an NVIDIA Quadro T2000 graphics 
card and a Intel Core i7-9850H CPU. Times are averaged over~2000 experiments for each scenario.

In particular, in Table~\ref{fig:dense_a} we 
evaluate a scenario of maximum size compatible with our
hardware, corresponding to a Dense layer of size 
$12800 \times 12800$ ($163,852,800$ parameters), and compare
it with several different convolutional layers with the same
total amount of FLOPs. 
The dense layer takes about~$6.4$ milliseconds (ms), 
while a unary convolution with $C_{in} = C_{out} = 3200$ on an input of spatial dimension 
$4\times 4$ just takes~$0.46$~ms, approximately~$16$ times faster.

In Figure~\ref{fig:dense_b}, we repeat the same experiment, varying
the total amount of flops with powers of~$2$. For the dense layer
we go from dimension $100\times 100$ to dimension 
$(100 \times 2^7) \times (100 \times 2^7)$.

In the following experiments, we keep the number of FLOPs 
constant while we increase some dimensions and proportionally decrease others.
If~\eqref{eq:conv_flops} had a good correlation with time, we should
observe straight horizontal lines.

\begin{figure}[ht]
\begin{subfigure}{.5\textwidth}
  \centering
  \includegraphics[width=.95\linewidth]{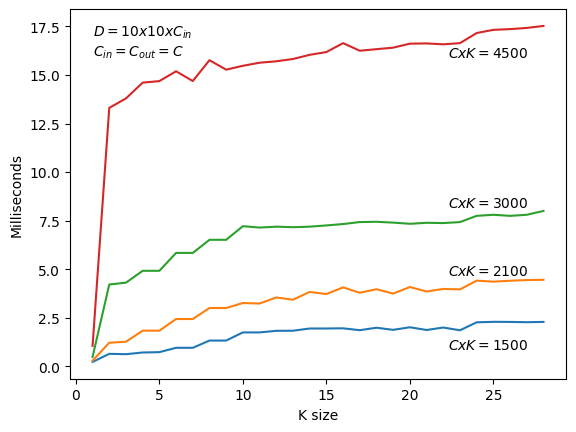}
  \caption{Increasing K, decreasing C}
  \label{fig:convsub_a}
\end{subfigure}%
\begin{subfigure}{.5\textwidth}
  \centering
  \includegraphics[width=.95\linewidth]{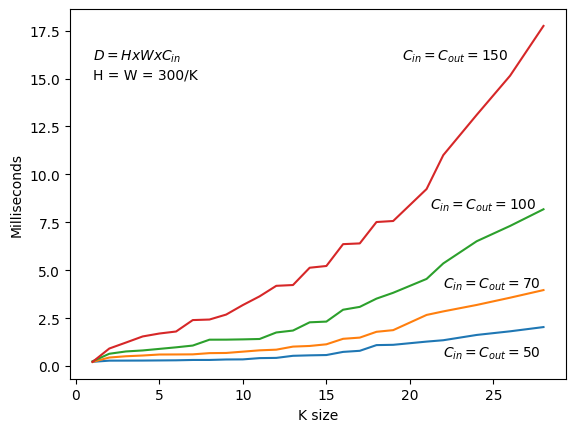}
  \caption{Increasing K, decreasing $W,H$}
  \label{fig:convsub_b}
\end{subfigure}

\begin{subfigure}{.5\textwidth}
  \centering
  \includegraphics[width=.95\linewidth]{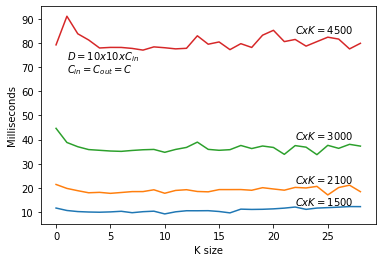}
  \caption{Same as (a) on a CPU}
  \label{fig:cpu_plot_a}
\end{subfigure}%
\begin{subfigure}{.5\textwidth}
  \centering
  \includegraphics[width=.95\linewidth]{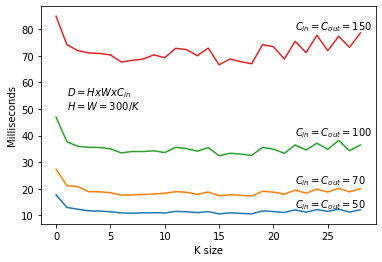}
  \caption{Same as (b) on a CPU}
  \label{fig:cpu_plot_b}
\end{subfigure}
\caption{Execution time vs different input dimensions, keeping
the number of FLOPs constant. In plot~(a) we increase~$K$ and 
proportionally decrease $C_\textit{in}$ and~$C_\textit{out}$.
In plot~(b) we increase~$K$ and proportionally decrease~$W$ and~$H$.
We would expect {\em constant lines}, but this is not the case.
In plots~(c) and~(d) we repeat the experiment on a (single core) CPU, instead of a GPU.}
\label{fig:conv_bad}
\end{figure}

In all experiments, we consider four different amounts
of FLOPs identified by different colors: $2025 \times 10^6$ (red line in Figure~\ref{fig:conv_bad}), $900 \times 10^6$ (green line), $490 \times 10^6$ (orange line) and $225 \times 10^6$
(blue line). We progressively increase~$K$ from~$1$ to~$30$. 
In the first experiment, we compensate it by enlarging the input and
output dimension of channels ($C_{in}$ and $C_{out}$), keeping a constant (small) spatial dimension $10 \times 10$.

In the second test we compensate the growing convolutions by 
reducing the spatial dimensions, starting from an initial 
dimension of $300 \times 300$. Channels are constant, in this case.
Result are reported in Figure~\ref{fig:conv_bad}. 

In the case of the first experiment (Figure~\ref{fig:convsub_a}), apart from the exceptional
performance of $1 \times 1$ convolutions already discussed in~\cite{trap_of_flops}, we observe the 
expected constant behavior. However, we have a completely different result in the
case of the second experiment (Figure~\ref{fig:convsub_b}). Here the execution time 
increases with the kernel dimension, possibly at a quadratic rate;
this growth should have been compensated by the simultaneous decrease
along both spatial dimensions, but clearly this is not the case.

By comparing the results of the two experiments, we can draw
another conclusion. Remember that the number of FLOPs along 
lines of the same color is the same; therefore,
the nonlinear behaviour in Figure~\ref{fig:convsub_b} is not due to an 
{\em overhead} but, on the contrary, there is 
an important {\em speed up} of the computation of increasing relevance for small kernels. 
In other words, the formula computing FLOPs is {\em overestimating} the total number of operations, 
presumably because it does not take into consideration the fact that convolutions can be {\em easily 
parallelized} along spatial dimensions (but not quite so along kernel dimensions).

The goal of the work is to derive a simple correction to the formula for computing FLOPs
explaining the observed behaviours. The correction might depend
on the specific hardware, but it should be possible to evaluate
the relevant parameters in a simple way. 

\section{$\alpha$-FLOPs}\label{sec:alpha_flops}

In this section we introduce our correction to the formula for computing FLOPs, that we call 
$\alpha$-FLOPs. Instead of FLOPs, that count the total number of floating point operations, 
$\alpha$-FLOPs provide an estimation of the \say{perceived} FLOPs, that are less than FLOPS due to 
parallelism. The crucial idea is that when we run in parallel an algorithm with a multidimensional 
input there is no reason to suppose that the total number of operations have similar latency along 
different dimensional axes. 
Our proposal is to adjust the formula for computing FLOPs by multiplying it by the following scaling factor:

\begin{equation}\label{eq:correction}
    \alpha_K(S)=\left(\frac{S_K + \beta_K(S-S_K)}{S}\right)^{\gamma_K}
\end{equation}

\noindent where $S = W\times H$, and $0 < \beta_K \ll 1, 0 < \gamma_K \le 1$, and $1 \le S_K \le S$ ($S_1 =1$) are parameters (moderately) depending from~$K$.
We call $\alpha$-FLOPs the correction to the usual formula for FLOPs by the previous factor. 

The parameters~$\beta_K$ and~$\gamma_K$ can be easily evaluated by regression on a given GPU/TPU. Although they are hardware dependent, some preliminary investigations seem to suggest that fluctuations are smaller than expected.

For the purposes of this article, using an Invida Quadro T2000 GPU we obtained good predictions just distinguishing two cases: $K=1$ and~$K>1$. For $K=1$, $\beta_K = 0.02$  and $\gamma_K = .99$; for $K>1$, $\beta_K = 0.001$ and $\gamma_K = .56$.

Before discussing the main properties of $\alpha_K(S)$, let us have a
look at the prediction of the execution time (dashed line) for the problematic experiments shown above. More examples will be presented in Section~\ref{sec:more_experiments}. 

\begin{figure}[ht]
\begin{subfigure}{.5\textwidth}
\begin{center}
\begin{tabular}{|c|c|c|}\hline
\multicolumn{3}{|c|}{327.68 M FLOPs}\\\hline\hline
\multicolumn{3}{|c|}{\bf Convolutional layers}\\\hline
config. & time & predicted\\\hline
 (1) & 6.392 & 6.154 \\\hline
 (2) & 3.224 & 3.351 \\\hline
 (3) & 1.626 & 1.847 \\\hline  
 (4) & 0.454 & 0.611 \\\hline
\end{tabular}
\caption{}
\end{center}
\end{subfigure}
\begin{subfigure}{.5\textwidth}
\includegraphics[width=1\textwidth]{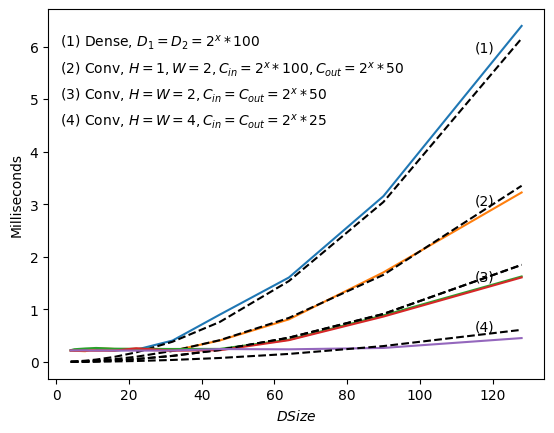}
\caption{\label{fig:plot6_b}}
\end{subfigure}
\caption{\label{fig:dense_vs_conv_predict}Predicted execution time
by means of $\alpha$-FLOPs for the same convolutional 
configurations of Figure~\ref{fig:dense_vs_conv}; in (b) predictions are depicted as dashed lines.}.
\end{figure}

The experiment in Figure~\ref{fig:dense_vs_conv} is replicated, with the time predicted
by means of $\alpha$-FLOPs, in Figure~\ref{fig:plot6_b}. In the Table on the left, we give the 
computed and predicted times for the convolutional configurations (1-4) with~327.68M FLOPs.

\begin{figure}[ht]
\begin{subfigure}{.5\textwidth}
  \centering
  \includegraphics[width=.95\linewidth]{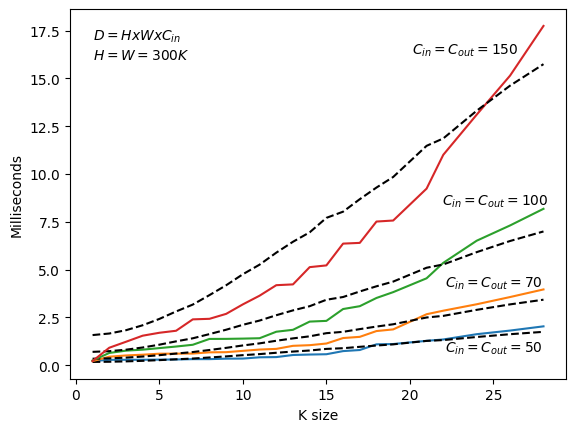}
  \caption{Increasing $K$, decreasing $C$}
  \label{fig:plot1}
\end{subfigure}
\begin{subfigure}{.5\textwidth}
  \centering
  \includegraphics[width=.95\linewidth]{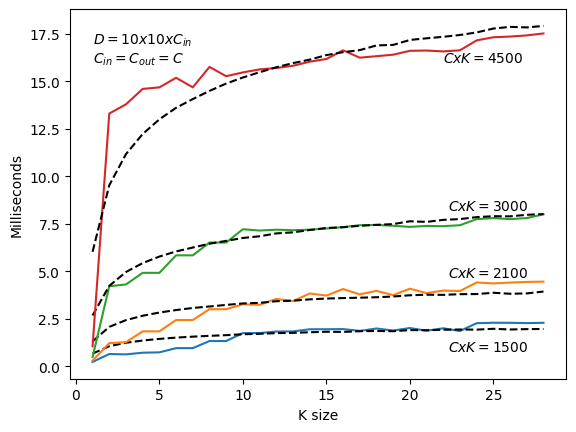}
  \caption{Increasing $K$, decreasing $W,H$}
  \label{fig:plot5}
\end{subfigure}
\caption{\label{fig:plot1-5}Predicted execution time
by means of $\alpha$-FLOPS, depicted as dashed lines, for the same convolutional 
configurations of Figure~\ref{fig:conv_bad}}
\end{figure}

Similarly, in Figure~\ref{fig:plot1-5} we show the predicted execution 
time for the experiments of Figure~\ref{fig:conv_bad}.

\subsection{Main properties of the $\alpha$-correction}
Before discussing our intuition behind~\eqref{eq:correction}, let us point out some of its 
distinctive properties. First of all the equation can be rewritten in the following, 
possibly more readable form:

\begin{equation}\label{eq:correction2}
\alpha_K(S) = \left((1-\beta_K)\times \frac{S_K}{S} + \beta_K\right)^{\gamma_K}
\end{equation}

From that, the following properties can be observed:

\begin{enumerate}

\item $\alpha_K(S) < 1$ for any~$K$ and~$S$. This is evident, given the constraint $S_k < S$.

\item If $\beta_K = 1$, then $\alpha_K(S) = 1$, independently from~$\gamma_K$ and~$S$. 
In this case, we recover the original expression for FLOPs, that is hence a subcase with no 
additional speedup.

\item $\alpha_1(1) = 1$ independently from $\beta_K$ and $\gamma_K$. This is due to the fact 
that $S_1=1=S$. The case $S=1, K=1$ is important since, as discussed at the 
beginning of Section~\ref{sec:comp-flops}, it gives the relation between convolutional
and dense layers, and for a dense layer we want no correction.
Moreover, the fact that the fundamental equation $\alpha_1(1) = 1$ holds independently 
from $\beta$ and $\gamma$ improves the stability of the property.

\item The formula with $\gamma=1$ already gives reasonable approximations. However, it tends to 
underestimate the execution time for large~$S$, in a more sensible way for increasing values
of~$K$. By raising the correction to a power smaller than 1 we mitigate this phenomenon 
without giving up the appealing properties provided by $\beta$.

\item The parameter~$S_K$ increases slowly with~$K$. 
The point is to
take into account a plausible overhead for growing dimensions of the kernel, especially when passing from $K=1$ 
to $K>1$. This constant can be possibly understood as a minimum expected spatial dimension for kernels larger 
than 1. It does not make much sense to apply a kernel of dimension $3\times 3$, on an input of dimension 
$1\times 1$, and it is hard to believe that reducing the dimension of the input below the dimension of the 
kernel may result in any speedup. However, fixing $S_K = K$ does not seem to be the right solution.

\end{enumerate}

\subsection{Rationale}
We now provide a possible explanation for the $\alpha$-FLOP formula~\eqref{eq:correction2}.
Let us consider a computational task requiring a given amount of work~$W$.
Let $\beta \in [0, 1]$ be the fraction of that work that can be executed in parallel; therefore,
the sequential portion of the task is $(1-\beta)W$.
Let us scale the problem by a factor $N > 1$; in a purely sequential framework, the
amount of work would become $NW$. However, Gustafson's law~\cite{Gustafson88}
suggests that when we scale the size of a parallel task, \emph{the sequential part
tend to remain the same}. This means that the amount of work \emph{actually} done
by the parallel program is $(1-\beta)W + \beta NW$. 
The ratio between the \emph{actual} amount of work done by the parallel version
versus the \emph{expected} amount of work done by the serial version is:

\begin{equation}\label{eq:backbone}
\frac{(1-\beta)W + \beta N W}{NW} = \frac{1-\beta}{N} + \beta
\end{equation}

where we readily recognize the backbone of equation~\eqref{eq:correction2}. We already discussed the small 
adjustments we had to do to this formula to fit it to the empirical observations.

Gustafson's law describes the theoretical speedup of a parallel task in terms of growing resources,  on the 
reasonable assumption that programmers tend to set the problem size to fully exploit the available 
computing power. Gustafson's law was meant to address a shortcoming of a similar law formulated by 
Amdahl~\cite{amdahl}, that on the other hand assumes a fixed workload that does not 
depend on the number of execution units used.

In our case, computational resources are fixed and we focus on different input dimensions. Our 
assumption is that suitable programs and load balancing techniques will optimize the use of 
resources, eventually resulting in different speedups along different spatial dimensions.

\section{Additional experimental results}\label{sec:more_experiments}

We conducted several experiments to assess the rate of grow of the execution time along different 
input dimensions. Data providing the base for this paper are available on~\href{https://github.com/asperti/alpha_flops_dataset}{Github} (\url{https://github.com/asperti/alpha_flops_dataset)}, together with analysis tools and 
plotting facilities, including the predictions by means of $\alpha$-FLOPs. 
Additional data are currently being collected.

All experiments discussed in this Section involve convolutions where we progressively increase the dimension of a 
specific input axis~$x$, keeping a constant dimension for the others axes~$X_c$. For each experiment, we draw the 
execution time for three different dimensions of an auxiliary axis $x_\textit{aux}$ in~$X_c$. We found this 
more understandable than plotting three dimensional surfaces, that would be quite difficult to draw and 
decipher.

\begin{figure}[ht]
\begin{subfigure}{.5\textwidth}
  \centering
  \includegraphics[width=.95\linewidth]{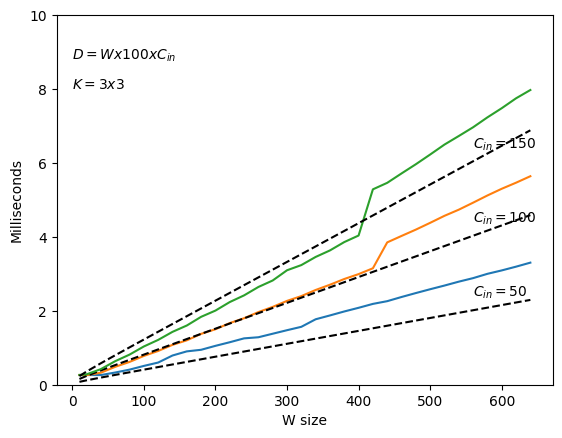}
  \caption{Increasing $W$ for different values of $C_\textit{in}$ and a kernel of dimension 3x3}
  \label{fig:plot2_K3}
\end{subfigure}
\begin{subfigure}{.5\textwidth}
  \centering
  \includegraphics[width=.95\linewidth]{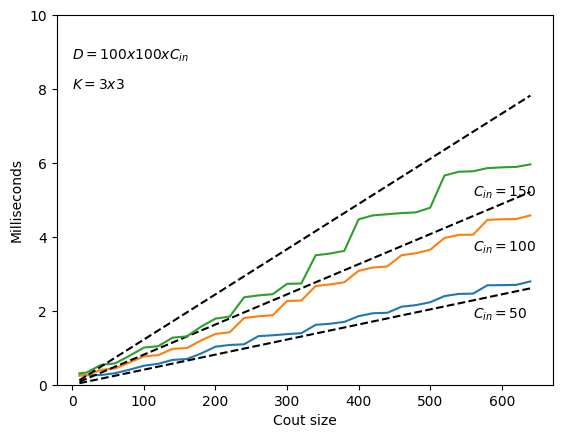}
  \caption{Increasing $C_\textit{out}$ for different values of $C_\textit{in} and a kernel of dimension 3x3$}
  \label{fig:plot3_K3}
\end{subfigure}
\caption{\label{fig:plots1}Execution time and predictions
by means of $\alpha$-FLOPs (dashed lines)}
\end{figure}

\begin{figure}[ht]
\begin{subfigure}{.5\textwidth}
  \centering
  \includegraphics[width=.95\linewidth]{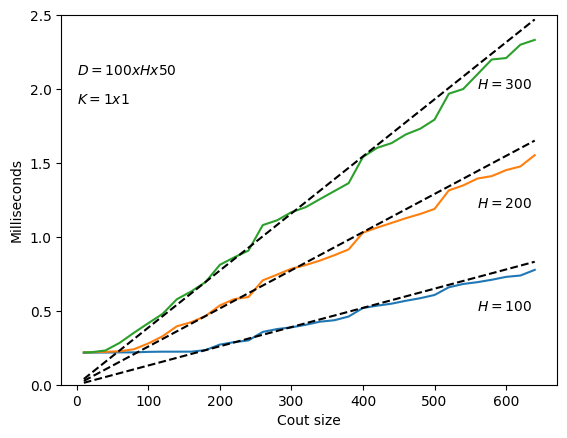}
  \caption{Increasing $C_\textit{out}$ for different values of H, with a kernel of dimension 1x1}
  \label{fig:plot4}
\end{subfigure}
\begin{subfigure}{.5\textwidth}
  \centering
  \includegraphics[width=.95\linewidth]{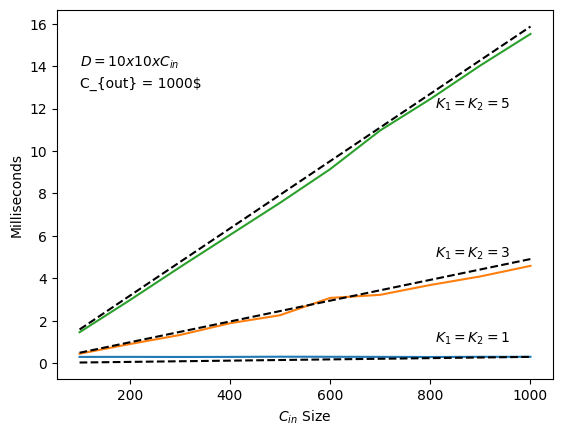}
  \caption{Increasing $C_\textit{in}$ for different kernels 1x1, 3x3 and 5x5}
  \label{fig:plot11}
\end{subfigure}
\caption{\label{fig:plots2}Execution time and predictions
by means of $\alpha$-FLOPS (dashed lines)}
\label{fig:vae}
\end{figure}

In the case of the plot in Figure~\ref{fig:plot2_K3}, $x$ is~$W$, 
and $x_\textit{aux}$ is $C_\textit{in}$; $H=100$ and the Kernel dimension is $3\times 3$. 
For Figure~\ref{fig:plot3_K3}, $x$ is $C_\textit{out}$,  
and $x_\textit{aux}$ is $C_\textit{in}$; $H=W=100$ and the kernel dimension is
$3\times 3$. For Figure~\ref{fig:plot4}, $x$ is $C_\textit{out}$,  
and $x_\textit{aux}$ is $H$; $C_\textit{in}=50$ and the kernel dimension is
$1\times 1$. Finally, for Figure~\ref{fig:plot11}, $x$ is $C_\textit{in}$, 
and $x_\textit{aux}$ is $K$; $H=W=10$ and $C_\textit{out}=1000$.

\subsection{Dense layers vs batchsize}
We already observed that a dense layer can be assimilated to
a convolutional layer with kernel $1 \times 1$ and spatial dimension~$1$.
In this perspective, it is plausible to conjecture that the batchsize can be assimilated to a spatial 
dimension. Indeed, the general wisdom across Deep Learning researchers and practitioners
is that, for making predictions over a large set of data --
e.g., over the full training or test set -- it is convenient to work
with a batchsize as large as possible, compatibly with
the resource constraints of the underlying hardware, e.g., memory. This has no
justification in terms of FLOPs, since the total number
of operations is always the same; however, using a large
batchsize is much more efficient. 

\begin{tabular}{cc}
\hspace{-.54cm}\begin{minipage}{.45\textwidth}
To test this behaviour, we take large dense layers (at the limit
of our hardware capacity), and then apply them to inputs
with increasing batch size.

The results are summarized in Figure~\ref{fig:dense}.
Under the FLOPs assumption, the lines should be straight lines
{\em departing from the origin}. In terms of $\alpha$-FLOPs we start
with the cost relative to batchsize~$1$, and then slowly grow
along the batchsize dimension, reflecting the experimental
behaviour. 

\end{minipage}&
\hspace{.4cm}\begin{minipage}{.5\textwidth}
\includegraphics[width=.95\linewidth]{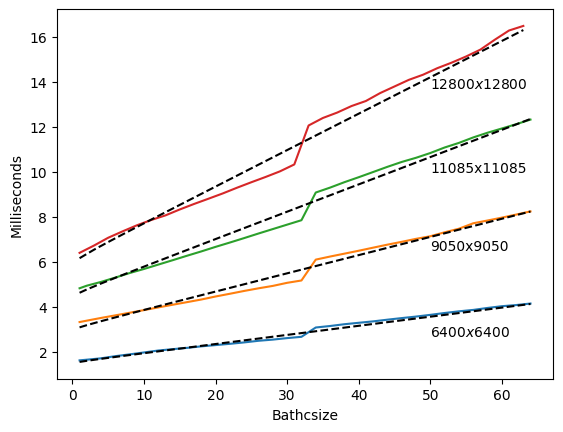}
\captionof{figure}{Computational time for dense layer increasing the batchsize. 
The jump between 32 and 33 is probably due to some
discretization in the software.}\bigskip
\label{fig:dense}
\end{minipage}
\end{tabular}




\section{Conclusions}\label{Sec:conclusions}
In this paper we introduced the notion of $\alpha$-FLOPs that
is meant to provide a simple numerical correction to the
mismatch between FLOPs and execution time 
in case of hardware equipped with massively parallel processing 
units like~GPUs or~TPUs. Since this kind of hardware is the norm
for~AI applications based on Deep Neural Networks, $\alpha$-FLOPS
may become an important tool to compare the efficiency of 
different networks on a given hardware. 

The definition of $\alpha$-FLOPs is based on the crucial
observation that, in case of an input with multiple dimensions,
the computational speedup offered by parallelism is typically far from
uniform along the different axes.
In particular, we provided extensive empirical evidence that growing spatial (and batchsize) 
dimensions in convolutional layers has less impact than growing different dimensions. 
The idea of dissecting the cost along the different input dimensions was inspired by recent investigations of the first author on computational complexity over finite types \cite{Asperti15}. 

The notion of $\alpha$-FLOPs lays between the number of parameters of the layer, and the traditional notion of FLOPs; in a sense, it can be  understood as a revaluation of the former as a measure of cost: if it is true that, in the case of convolutions, the  number of parameters does not take into account the actual cost of the convolution, the traditional notion of FLOPs seems to largely overestimate it.

Much work is still ahead. On the experimental side, we are currently collecting more data, on architectures with different computing capabilities. On the theoretical side,
it would be interesting to provide a low-level algorithmic justification of $\alpha$-FLOPs. The formula 
itself, that was derived empirically, can be eventually fine-tuned and possibly improved, both in view of 
additional observations, and of a better understanding of the phenomenon. In particular, we mostly focused on 
the spatial dimension, since it is the axis most affected by parallelism, but the dependency along different 
axes does eventually deserve additional investigation.

In this article, we mostly focused on convolutional and dense layers, since they are the most computationally intensive layers in Neural Networks. Extending the work to additional layers, or more sophisticated
forms on convolutions, like Depth-Separable Convolutions, is another major research direction.

\bibliographystyle{plain}
\bibliography{machine.bib,variational.bib}

\end{document}